\documentclass[letterpaper, 10 pt, conference]{ieeeconf}  

\IEEEoverridecommandlockouts                              

\usepackage{graphics} 
\usepackage{epsfig} 
\usepackage{times} 
\usepackage{amsmath} 

\usepackage{amsfonts}       
\usepackage{graphicx}
\usepackage{bm}
\usepackage{tabularx}
\usepackage{floatrow}
\usepackage[utf8]{inputenc}
\usepackage{amsmath}
\usepackage[utf8]{inputenc} 
\usepackage[T1]{fontenc}    
\usepackage[mathscr]{eucal}
\usepackage{subfig}
\usepackage{caption}
\usepackage{mcite}
\usepackage{epsfig}
\usepackage{makecell}
\usepackage{multirow}
\usepackage{hyperref}

\usepackage{tabularx,ragged2e,array}

\newcommand{\prl}[1]{\left(#1\right)}

\title{\LARGE \bf
Learning Sample-Efficient Target Reaching for Mobile Robots}
\author{Arbaaz Khan$^{1}$, Vijay Kumar$^{1,2}$, Alejandro Ribeiro$^{2}$
\thanks{$^{1}$GRASP Lab, University of Pennsylvania, Philadelphia, PA 19104, USA,  arbaazk@seas.upenn.edu. $^{2}$Department of Electrical and Systems Engineering, University of Pennsylvania, Philadelphia, PA 19104, \{aribeiro, kumar\}@seas.upenn.edu.}}

\begin{document}
\maketitle
\thispagestyle{empty}
\pagestyle{empty}

\begin{abstract}
In this paper, we propose a novel architecture and a self-supervised policy gradient algorithm, which employs unsupervised auxiliary tasks to enable a mobile robot to learn how to navigate to a given goal. The dependency on the global information is eliminated by providing only sparse range-finder measurements to the robot. The partially observable planning problem is addressed by splitting it into a hierarchical process. We use convolutional networks to plan locally, and a differentiable memory to provide information about past time steps in the trajectory. These modules, combined in our network architecture, produce globally consistent plans. The sparse reward problem is mitigated by our modified policy gradient algorithm. We model the robots uncertainty with unsupervised tasks to force exploration. The novel architecture we propose with the modified version of the policy gradient algorithm allows our robot to reach the goal in a sample efficient manner, which is orders of magnitude faster than the current state of the art policy gradient algorithm. 
Simulation and experimental results are provided to validate the proposed approach.
\end{abstract}
\section{INTRODUCTION}

When a robot is deployed in an unknown environment where it is tasked to identify a specific object, we need access to algorithms that enable a robot to reach a target without prior knowledge of the environment. This canonical problem in robotics is challenging because of the difficulty to hand engineer feature detection or mapping techniques that work in the full variety of environments the robot may encounter. This motivates the use of learning techniques to design policies that map sensor inputs to control outputs. In particular, deep reinforcement learning has emerged as a successful approach to design such policies for complex tasks where it is difficult or impossible to engineer features. Example problems where deep reinforcement learning has proven successful include maximizing video game score~\cite{dqn}, learning how to grasp objects with a robot arm~\cite{visuomotor}, or learning obstacle avoidance policies for quadrotors~\cite{zhang}. In this work, we use deep reinforcement learning to solve the problem of target reaching for a mobile robot in an unknown environment where the only available information to the robot are the readings of an on board range finder. 

Deep learning has been applied to trajectory planning in environments that can be fully~\cite{vin} or partially~\cite{ohpaper, chen2017neural, parisotto2017neural, gupta2017cognitive} observable. These works differ in training methodologies which fall in two main categories: (i) Training with supervised learning where an expert provides optimal policies \cite{chen2017neural, gupta2017cognitive}. (ii) Training with reinforcement learning where the robot gets rewards for exploring the environment and uses the received rewards to update its belief state \cite{ohpaper, parisotto2017neural}. Our focus is on the latter training paradigm. We want to design algorithms where the robot starts exploring under a random policy and collects rewards by interacting with the environment. Using these rewards, the robot updates the policy without supervision so that it eventually learns a good policy to explore the environment \cite{sutton1998reinforcement}. 

\begin{figure}[t!]
  \centering
  \includegraphics[scale = 0.26]{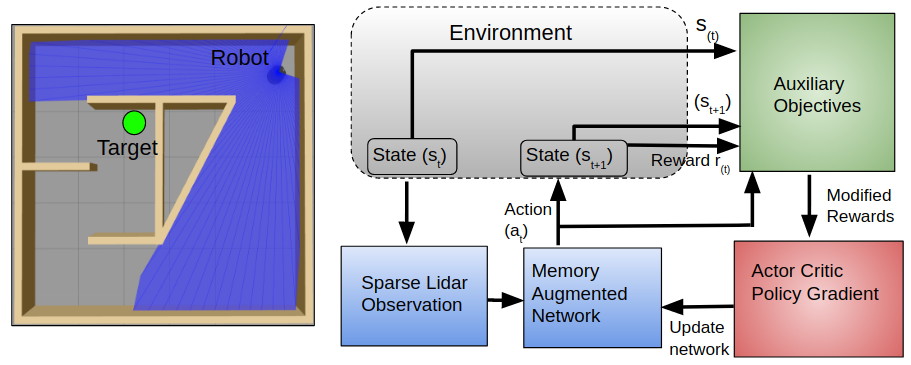}
  \caption{Observation at state $s_{t}$ are input to the memory augmented network as a sparse lidar scan. The network outputs an action $a_{t}$ which results in a new state $s_{t+1}$ and a reward $r_{t}$. $s_t$, $a_t$, $r_t$ are used to model the uncertainty of the robot using auxiliary objectives. This is used as a pseudo reward that is added to the external reward $r_t$. Modified reward is then used in the policy gradient step to update the weights of the network.\label{fig:page1}}
\end{figure}

Such a training paradigm is complicated by the fact that meaningful reward signals are collected far into the future. Indeed, the robot collects no reward until it reaches the goal where it collects a reward that is discounted to the present time. Since the target is typically several time steps away, the current reward signal provides minimal information to guide the learning loop. This is the well known credit assignment problem in reinforcement learning which in the context of trajectory planning has motivated the use of environmental markers to guide the robot towards the goal \cite{ohpaper, parisotto2017neural}. It is important to observe, however, that the use of markers is somewhat inherently incompatible with the idea of self supervised learning. Another not unrelated problem when applying deep reinforcement learning for trajectory planning is the lack of memory that is inherent to Markov decision processes. This precludes successful navigation because optimal actions in planning problems tend to be sequential in nature and necessitate long term memory representations of states already visited~\cite{zhangmemory}.

Our goal in this paper is to propose a deep reinforcement learning architecture that overcomes the twin problems of long term memory and credit assignment. Our specific contributions are: 

\hangindent=10pt {\bf Memory augmentation.} We propose a novel memory augmented mapless planner. The input to the planner is a lidar scan and no other information. However, the memory augmentation helps to model the sequential dependency of actions and enables learning of large-reward trajectories.

\hangindent=10pt {\bf Learning with auxiliary tasks.} The planner is trained end-to-end by policy gradient algorithms that are partnered with unsupervised auxiliary tasks. The auxiliary tasks produce estimates of states, actions, and rewards that are used to alter the reward structure of the Markov decision process. This modified rewards help a non holonomic robot to learn policies that enable it to reach the goal in the absence of meaningful reward signals from the environment. 

\noindent We begin the paper by formulating trajectory planning in unknown environments as a Markov decision process (Section II) and introduce policy gradient and actor critic loss formulations (Section II.A). To force the robot to explore the environment in the absence of meaningful reward signals from the environment, we create auxiliary tasks to model the uncertainty of the robot (Section II.B). We do this by training neural networks to predict state uncertainty, reward uncertainty and action uncertainty. These networks take in the current state $s_t$ and action $a_t$ and output a prediction for the next state $\hat{s}_{t+1}$, an estimate of the improved action $\hat{a}_t$, and an estimate of the reward $\hat{r}_t$. If these predictions differ significantly from the observed state ${s}_{t+1}$, the action ${a}_t$, or the actual reward ${r}_t$, this is interpreted as a signal to force the robot to explore the environment. 

In this work, the thesis put forward is that learning how to navigate to a given goal in partially observable unknown environments with sparse rewards consists of two key steps: planning and exploration. We split the planning problem into two levels of hierarchy. At a lower level the network uses an explicit planning module and computes optimal policies using a feature rich representation of the locally observed environment. This local policy along with a sparse representation of the partially observed environment is part of the optimal solution in the global environment. To compute optimal global policies from our local policies, we augment our local planning module with an external memory scheme. This memory network was trained by supervision provided by expert policies. We extend our related earlier work on memory augmented network~\cite{macn} for planning to a self supervised training scheme where in addition to learning how to plan, it also learns how to explore the environment by itself. 
We show our proposed memory augmented network with auxiliary awards is able to reach a target region in a partially observable environment in both simulation and when transferred to the real world. Additionally, when compared with relevant baselines, it offers a considerable speedup in convergence.

\section{Deep Reinforcement Learning with Auxiliary Objectives}

Consider an robot with state $s_t \in \mathcal{S}$ at discrete time $t$. Let the states $\mathcal{S}$ be a discrete set $[s_1, s_2, \ldots, s_n]$. For a given action $a_t \in \mathcal{A}$, the robot evolves according to known deterministic dynamics: $s_{t+1} = f(s_t,a_t)$. The robot operates in an unknown environment and must remain safe by avoiding collisions. Let $m \in \{-1,0\}^n$ be a \textit{hidden} labeling of the states into free $(0)$ and occupied $(-1)$. The robot has access to a sensor that reveals the labeling of nearby states through an observations $z_t = H(s_t)m$, where $H(s) \in \mathbb{R}^{n \times n}$ captures the local field of view of the robot at state $s$. The local observation consists of readings from a range finder. The observation $z_t$ contains zeros for unobservable states. Note that $m$ and $z_t$ are $n \times 1$ vectors and can be indexed by the state $s_t$.  The robot's task is to reach a goal region $\mathcal{S}^{\text{goal}} \subset \mathcal{S}$, which is assumed obstacle-free, i.e., $m[s] = 0$ for all $s \in \mathcal{S}^{\text{goal}}$. The information available to the robot at time $t$ to compute its action $a_t$ is $h_t := \prl{s_{0:t}, z_{0:t}, a_{0:t-1}} \in \mathcal{H}$, where $\mathcal{H}$ is the set of possible sequences of observations, states, and actions. \textit{The partially observable target reaching problem can then be stated as follows} :

{\textbf{Problem 1:}}
Given an initial state $s_0 \in \mathcal{S}$ with $m[s_0] = 0$ (obstacle-free) and a goal region $\mathcal{S}^{\text{goal}}$, find a function $\mu : \mathcal{S} \rightarrow \mathcal{A}$ such that applying the actions $a_t := \mu(s_t)$ results in a sequence of states $s_0,s_1,\ldots,s_T$ satisfying $s_T \in \mathcal{S}^{\text{goal}}$ and $m[s_t] = 0$ for all $t = 0, \ldots, T$.  

Instead of trying to estimate the hidden labeling $m$ using a mapping approach, our goal is to learn a policy $\mu$ that maps the sequence of sensor observations $z_0, z_1, \ldots z_T$ directly to actions for the robot. The partial observability requires an explicit consideration of \textit{memory} in order to learn $\mu$ successfully. A partially observable problem can be represented via a Markov Decision Process (MDP) over the history space $\mathcal{H}$. More precisely, we consider a finite-horizon discounted MDP defined by $\mathcal{M}(\mathcal{H},\mathcal{A},\mathcal{T},r,\gamma)$, where $\gamma \in (0,1]$ is a discount factor, $\mathcal{T}: \mathcal{H} \times \mathcal{A} \rightarrow \mathcal{H}$ is a deterministic transition function, and $r: \mathcal{H} \rightarrow \mathbb{R}$ is the reward function, defined as follows:
\begin{align*}
  \mathcal{T}(h_t,a_t) &= (h_t,s_{t+1} = f(s_t,a_t), z_{t+1} = H(s_{t+1})m, a_t)\\
  r(h_t,a_t) &= z_t[s_t]
\end{align*}
The reward function definition stipulates that the reward of a state $s$ can be measured only after its occupancy state has been observed.

Given observations $z_{0:t}$, we can obtain an estimate $\hat{m} = \max\{\sum_\tau z_\tau,-1\}$ of the map of the environment and use it to formulate a locally valid, fully-observable problem as the MDP $\mathcal{M}_t(\mathcal{S},\mathcal{A},f,r,\gamma)$ with transition function given by the robot dynamics $f$ and reward $r(s_t) := \hat{m}[s_t]$ given by the map estimate $\hat{m}$.

\subsection{Policy Gradients and Actor Critic}
Consider the locally valid fully observable MDP $\mathcal{M}_t(\mathcal{S},\mathcal{A},f,r,\gamma)$. The goal of any reinforcement learning algorithm is to find a policy $\pi(a_t|s_t;\theta)$ (where $\theta$ are the parameters of the policy) that maximizes the expected sum of rewards :
\begin{equation}
\max_{\theta}  \mathbb{E}_{\pi(a_t|s_t;\theta)}[\sum_{t} r_t]  \enspace
\end{equation}
Policy gradient methods look to maximize the reward by estimating the gradient and using it in a stochastic gradient ascent algorithm. A general form for the policy gradient can be given as:
\begin{equation}
\hat{g} = \hat{\mathbb{E}}_t[\nabla_{\theta}\text{log} \pi_{\theta}(a_t|s_t)\hat{A}_t ^{\pi}(s_t,a_t)]
\end{equation}
where the $\hat{A}_t ^{\pi}(s_t,a_t)$ represents the advantage function and is given as :
\begin{equation}
\hat{A}_t ^{\pi}(s_t,a_t) := Q^{\pi}(s_t,a_t) - V^{\pi}(s_t) 
\end{equation}
Further, $Q^{\pi}(s_t,a_t)$ represents the action value function (estimate of how good it is to take an action in a state) and $V^{\pi}(s_t)$ represents the state value function (estimate of how good it is to be in a state) and they are given as:
\begin{equation}
Q^{\pi}(s_t,a_t) := \mathbb{E}_{s_{t+1:\infty},a_{t+1:\infty}} \Big[ \sum_{l=0}^{\infty} r_{t+l} \Big] 
\end{equation}
\begin{equation}
V^{\pi}(s_t) := \mathbb{E}_{s_{t+1:\infty},a_{t:\infty}} \Big[ \sum_{l=0}^{\infty} r_{t+l} \Big] 
\end{equation}
The gradient $\hat{g}$ is estimated by differentiating the objective: 
\begin{equation}
L^{PG}(\theta)=\mathbb{E}[\text{log}(\pi_{\theta}(a_t|s_t)\hat{A}_t]
\end{equation}
This is known as the policy gradient algorithm. In practice, one does not know the advantage function beforehand and it must be estimated. Most implementations for computing advantage function estimators~\cite{mnih2016asynchronous} use a deep network to learn the state-value function $V(s_t)$. This learned value function is used to compute the policy as well as the advantage function. When using a network architecture that shares parameters between the policy function and the value function, we also need to include a value function error term in the loss function for the network. Additionally as shown in~\cite{mnih2016asynchronous} one can also add an entropy bonus to ensure sufficient exploration. Using these,  policy gradient is modified to produce the actor critic algorithm maximizes the following objective function:
\begin{equation}\label{eqn:ac}
L_{AC} = \hat{\mathbb{E}}_t[\text{log}\pi_{\theta}(a_t|s_t)\hat{A}_t - \alpha (V_{\theta}(s_t)-V_t^{targ})^2 + \beta \text{S}[\pi_{\theta}](s_t)]
\end{equation}
where $(V_{\theta}(s_t)-V_t^{targ})^2$ is the squared-error loss for the value function term, $[\pi_{\theta}](s_t)$ is the entropy term and $\alpha$ and $\beta$ are hyper parameters.

To remove any dependence on environment specific features, instead of using raw data from the environment for the auxiliary tasks, we use a encoding $\phi(s_t)$ that is learned by a deep network. This encoding is used for the actor-critic loss as well as the auxiliary objectives. Then Eqn \ref{eqn:acenc} can be rewritten as:
\begin{equation}\label{eqn:acenc}
\begin{split}
L^{enc}_{AC} = \hat{\mathbb{E}}_t[\text{log}\pi_{\theta}(a_t|\phi(s_t))\hat{A}_t - \alpha (V_{\theta}(\phi(s_t))-V_t^{targ})^2 + \\
\beta \text{S}[\pi_{\theta}]\phi(s_t)]
\end{split}
\end{equation}

\subsection{Modelling Robot Uncertainty with Auxiliary Objectives}

\begin{figure}[t!]
  \centering
  \includegraphics[scale = 0.375]{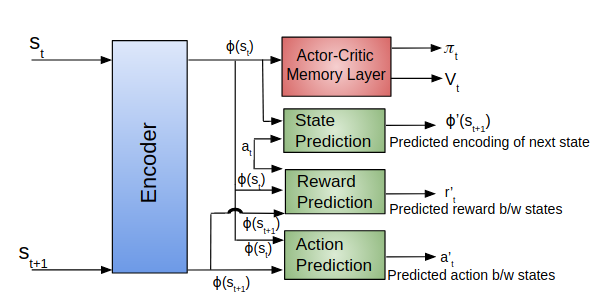}
  \caption{Instead of using direct raw observations, an encoding to represent the state is learned. This encoding is used to minimize the actor critic loss in Eqn. \ref{eqn:ac}. The same encoding is used as input to the networks that minimize the auxiliary losses defined in Eqn.\ref{eqn:sp} and Eqn.\ref{eqn:rp}. By sharing the encoding between the policy gradient and the auxiliary tasks the robot learns to balance improving its performance for the navigation task with improving performance for the auxiliary tasks. This leads to faster convergence.\label{fig:auxrewards}}
\end{figure}

In this paper, we propose auxiliary objectives that are suited to real world robotics tasks and can be easily adapted to most reinforcement learning problems for robots. Our proposed auxiliary objectives are unsupervised and do not need any re-engineering when going from one environment to another or from one kind of sensor to another. We use these auxiliary objectives as a small reward signal internally in the absence of any reward signal from the environment. In reinforcement learning literature these are called as pseudo-reward functions. Fig \ref{fig:auxrewards} gives an overview of the auxiliary objectives along with the actor-critic policy gradient.

\medskip\noindent{\textbf{Next State Prediction}}: The confidence of the robot in its learned transition model is given by a neural network with parameter $\theta_{SP}$. The neural network takes in as input the encoding of the current state $\phi(s_t)$, the action executed $a_t$ and outputs a prediction for the encoding of the next state $\hat{\phi}(s_{t+1})$. The loss function for this auxiliary task $L_{SP}$ and the pseudo reward $r_{SP}$ is then given as:
\begin{equation}\label{eqn:sp}
L_{SP}(\hat{\phi}(s_{t+1}),\phi(s_{t+1});\theta_{SP}) = \frac{1}{2}||\hat{\phi}(s_{t+1})-\phi(s_{t+1})||_2^2
\end{equation}
\begin{equation}
    r_{SP}= 
    \begin{cases}
      L_{SP}, & \text{if}\ -\eta_{SP} \leq L_{SP} \leq \eta_{SP} \\
      1.5, & \text{otherwise}
    \end{cases}
\end{equation}
The pseudo reward is bounded by $\eta_{SP}$ to ensure that the next state prediction does not overshadow the main task of reaching the goal, i.e $\sum_{t}r_{SP} \ll \sum_{t}r_{t}$. 

\medskip\noindent{\textbf{Reward Prediction}}:
For this task, another neural network with parameters $\theta_{RP}$ is used to model the robots confidence in its learned environment dynamics. The neural network takes in as input the encoding of the current state $\phi(s_{t})$, the action $a_t$ sampled from the current policy $\pi$ and the encoding of the resulting next state $\phi(s_{t+1})$ and outputs a prediction for the reward $\hat{r}_t$ received from the environment. To account for temporally delayed rewards, the sequence reward prediction  The loss function for this auxiliary task $L_{RP}$  and pseudo reward $r_{SP}$ is then given as:
\begin{equation}\label{eqn:rp}
L_{RP}(\hat{r}_{t},r_{t};\theta_{RP}) = \frac{1}{2}||\hat{r}_{t}-r_{t}||_2^2
\end{equation}
\begin{equation}
    r_{SP}= 
    \begin{cases}
      L_{RP}, & \text{if}\ -\eta_{RP} \leq L_{RP} \leq \eta_{RP} \\
      1.5, & \text{otherwise}
    \end{cases}
\end{equation}
Similar to the pseudo reward in the next state prediction, we also bound the pseudo reward by $\eta_{RP}$ in this task to ensure that the pseudo rewards from auxiliary reward prediction task does not overshadow rewards for the main task. 

\medskip\noindent{\textbf{Action Prediction}}:
Lastly, another neural network with parameters $\theta_{AP}$ is used to model the learned uncertainty in the robots control dynamics. Alternatively, this can also be thought of as learning the transition probability $f$ in the MDP $\mathcal{M}_t$. The neural network takes in as input the encoding of the current state $\phi(s_t)$ and the encoding of the next state $\phi(s_{t+1})$ and outputs a predicted control action $\hat{a}_t$. Since the action space is discrete (see experimental setup), the final layer of the actor-critic network outputs a softmax over all possible actions and gives probability of each action. Therefore, for action prediction the auxiliary loss $L_{AP}$ is setup as a cross-entropy classification across the discrete action classes. No pseudo reward from the action prediction is used to modify the global reward signal from the environment. 
Thus, the final loss function that is minimized can be written as: 
\begin{equation}
\min_{\theta, \theta_{RP},\theta_{SP},\theta_{AP}}L_{total} = L^{enc}_{AC} + \lambda_{SP} L_{SP} + \lambda_{RP} L_{RP} + \lambda_{AP} L_{AP} 
\end{equation}
where $ 0 \leq \lambda_{SP}$, $\lambda_{RP}$ $\lambda_{AP} \leq 1$ are hyperparameters that represent weighting on their respective loss functions and the objective now is to learn a policy $\pi(a|s_t)$ that maximizes the expected sum of modified rewards \footnote{Here $\theta$ represents the joint space of parameters $\theta, \theta_{RP},\theta_{SP},\theta_{AP}$}:
\begin{equation}
\max_{\theta}  \mathbb{E}_{\pi(a_t|s_t;\theta)}[\sum_{t} r_t + r_{SP} + r_{AP}]  \enspace
\end{equation}

The idea of incorporating auxiliary tasks into the reinforcement learning problem to promote faster convergence (or training) has only been recently explored. The authors in ~\cite{van2017hybrid},~\cite{mirowski},~\cite{actunreal} showed that using auxiliary objectives in navigation tasks speeds up training considerably. However, these results are on a simulated video game and use auxiliary objectives such as depth prediction and change in pixel values from frame to frame. These objectives work well in a video game environment, but do not scale well to a real world environment where depth measurements and color values are highly susceptible to noise.

\section{Memory Augmented Control Network for Actor-Critic}
In this section, we introduce the memory augmented network architecture.  We propose a memory augmented value iteration network (MACN) that split the planning problem in a partially observable environment into two steps. At a lower level, a value iteration network~\cite{vin} was used to compute local plans from the locally observed space and at a higher level a differentiable memory computed global consistent plans by using a history of these local plans and low level features from the observation space. 
In this section, we briefly describe the memory augmented value iteration networks and then describe how it is modified to learn policies from sensor input alone for the partially observable target reaching problem. 

\subsection{Value Iteration Networks and Differentiable Neural \enspace Memory}
Consider a finite horizon discounted MDP denoted by $\mathcal{M}(\mathcal{S}, \mathcal{A}, f, r, \gamma)$. The solution of such MDP is a policy $\pi(a|s)$ that obtains maximum rewards. The classical algorithm to solve an MDP is Value Iteration (VI)~\cite{sutton1998reinforcement}. The value of a state $V^{\pi}(s_t)$ is computed iteratively. 
The update rule for value iteration is given as:
\begin{equation}\label{eq:VI}
    V_{k+1}(s) =  \max_{a \in \mathcal A} [R(s, a) + \gamma \sum_{s' \in S} f(s' | s, a) V_k(s)]
\end{equation}
Additionally, the standard form for windowed convolution is
\begin{equation}\label{eq:winConv}
    V(x) = \sum_{k=x-w}^{x+w} V(k) f(k) 
\end{equation}
\begin{figure*}[t!]
  \includegraphics[width=\textwidth]{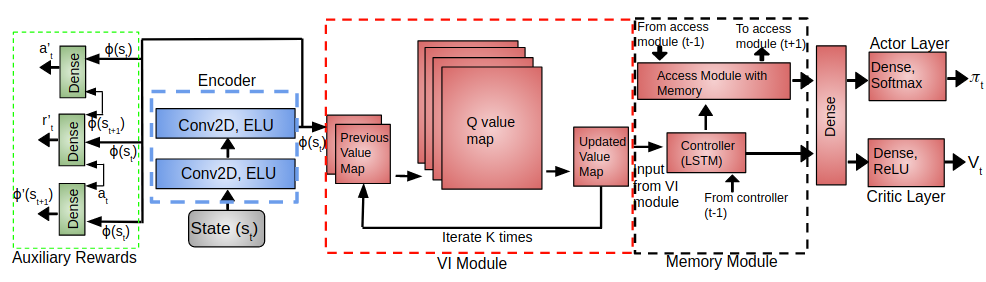}
  \caption{Memory Augmented Actor Critic with Auxiliary Reward Architecture for the Partially Observable Target Reaching Problem. Here dense implies fully connected layers.}\label{fig:arch}
\end{figure*}
Tamar et al.~\cite{vin} show that the summation in Eqn.~\eqref{eq:winConv} is analogous to $\sum_{s'} f(s'|s,a) V_k(s)$. When Eqn.~\eqref{eq:winConv} is stacked with reward, max pooled and repeated K times, it represents an approximation of the value iteration algorithm over K iterations. The memory augmented network scheme uses a value iteration network to compute a plan for the locally observed space.

The differentiable neural memory (DNC) introduced in~\cite{dnc} uses an external memory bank $M$ in conjunction with a recurrent neural network (RNN). The recurrent neural network learns to read, write and access specific locations in the external memory. The recurrent neural network learns to output read weights ( $w_{t}^R$) and write weights ($w^{W}_{t}$). These are used to selectively write to memory and to read from memory. \footnote{We refer the interested reader to the original Value Iteration Networks~\cite{vin} and the DNC~\cite{dnc} paper for a complete description.}
In our previous related work ~\cite{macn} we introduced a similar memory architecture for planning. However, a key difference between this paper and our earlier work is that in this paper, we learn our policies with reinforcement learning as compared to our older work where supervised learning was used. However, when~\cite{macn} is applied directly to a robotics navigation problem, it presents several challenges. It needs a perfect labeling of states and it does not take into account occlusion effects of sensors, presents significant challenges when scaling it up to large environments and lastly, it is setup to be trained by supervised learning and generating expert policies is often non trivial. 

\subsection{MACN for Self-Supervised Partially Observable Target Reaching}

\begin{figure}[h!]
  \centering  \includegraphics[scale = 0.1]{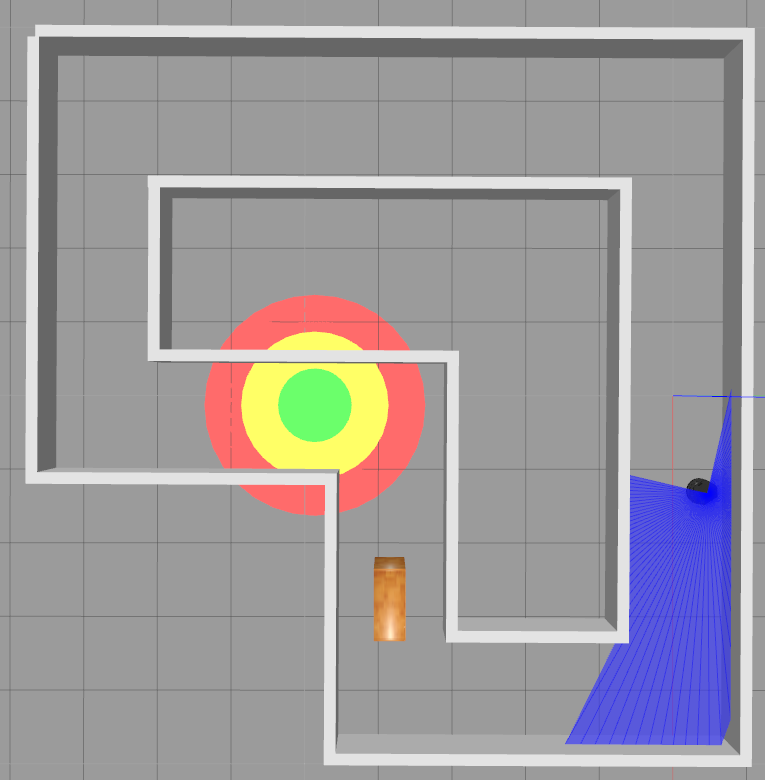}
  \caption{\textbf{Partially Observable Environment with Long Corridors} The robot only observes a small part of the environment through its sensor and using only this information, it must navigate to the goal region (bullseye). The brown bookshelf blocks the easy path of the robot. \label{fig:circuit}}
\end{figure}

Consider the world  in Fig \ref{fig:circuit}. The goal of the robot is to learn how to navigate to the goal region. Let this environment in Fig \ref{fig:circuit} be represented by a MDP $\mathcal{M}$. 
The key intuition behind designing the memory augmented architecture is that planning in $\mathcal{M}$ can be decomposed into two levels. At a lower level, planning is done in a local space within the boundaries of our locally observed environment space. Let this locally observed space be $z'$. As stated before in Section II, this observation can be formulated as a fully observable problem $\mathcal{M}_t(\mathcal{S},\mathcal{A},f,r,\gamma)$. It is possible to plan in $\mathcal{M}_t$ and calculate the optimal policy for this local space,  $\pi_{l}^*$ independent of previous observations. It is then possible to use any planning algorithm to calculate the optimal value function $V^*_{l}$ from the optimal policy $\pi_l^*$ in $z'$. Let $\Pi=[\pi_{l}^{1},\pi_{l}^{2},\pi_{l}^{3},\pi_{l}^{4},\dots,\pi_{l}^{n}]$ be the list of optimal policies calculated from such consecutive observation spaces [$z_0,z_1,\dots z_T$]. Given these two lists, it is possible to train a convolutional neural network with supervised learning.The network could then be used to compute a policy $\pi_{l}^{new}$ when a new observation $z^{new}$ is recorded.

This policy learned by the convolutional network is purely reactive as it is computed for the $z^{new}$  observation independent of the previous observations. Such an approach fails when there are local minima in the environment. In a 2D/3D world, these local minima could be long narrow tunnels culminating in dead ends. In the scenario shown in Fig \ref{fig:circuit} the environment has a long corridor that is blocked at one end. The environment is only partially observable and the robot has no prior knowledge about the structure of this corridor forcing it to explore the corridor all the way to the end. Further, when entering and exiting such a structure, the robot's observations are the same, i.e $z_1 = z_2$, but the optimal actions under the policies $\pi_{l}^{1}$ and $\pi_{l}^{2}$ (computed by the convolutional network) at these time steps are not the same, i.e $a_{\pi^1} \neq a_{\pi^2}$. To backtrack successfully from such a corridor, information about previously visited states is required, necessitating memory. 

To solve this problem, we propose using a differentiable memory to estimate the map of the environment $\hat{m}$. The controller in the memory network learns to selectively read and write information to the memory bank. When such a differentiable memory scheme is trained it is seen that it keeps track of important events/landmarks (in the case of corridor, this is the observation that the blocked end has been reached) in its memory state and discards redundant information. In theory one can use a CNN to extract features from the observation $z'$ and pass these features to the differentiable memory. Instead, we propose the use of a VI module~\cite{vin} that approximates the value iteration algorithm within the framework of a neural network to learn value maps from the local information. We hypothesize that using these value maps in the differential memory scheme 
provides us with better planning as compared to when only using features extracted from a CNN. This architecture is shown in Fig ~\ref{fig:arch}.

The VI module is setup to learn how to plan on the local observations $z$. The local value maps (which can be used to calculate local policies) are concatenated with a low level feature representation of the environment and sent to a controller network. The controller network interfaces with the memory through an access module (another network layer) and emits read heads, write heads and access heads. In addition, the controller network also performs its own computation for planning. The output from the controller network and the access module are concatenated and sent through a linear layer to produce an action. This entire architecture is then trained end to end. Thus, to summarize, the planning problem is solved by decomposing it into a two level problem. At a lower level a feature rich representation of the environment (obtained from the current observation) is used to generate local policies. At the next level, a representation of the histories that is learned and stored in the memory, and a sparse feature representation of the currently observed environment is used to generate a policy optimal in the global environment.

\section{Experiments}
\begin{figure*}[t!]
\centering
\includegraphics[scale = 0.5]{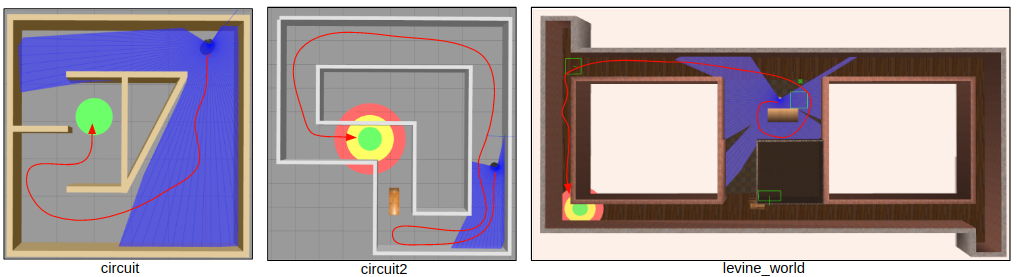}\quad

\medskip

\includegraphics[width=\textwidth]{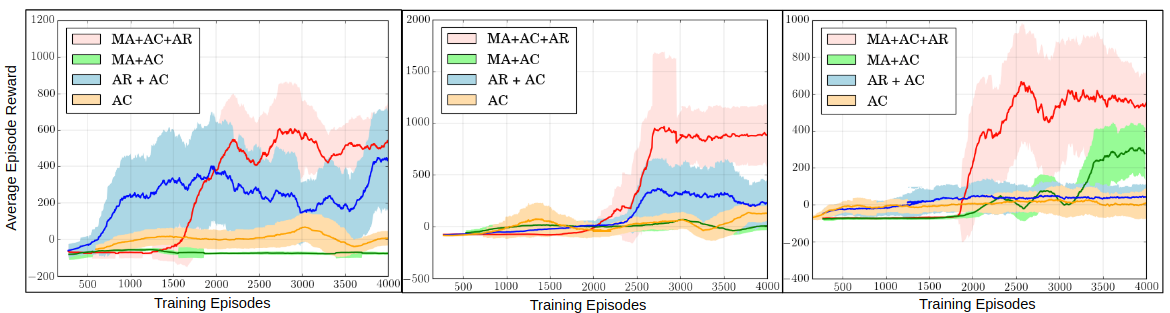}\quad

\caption{\textbf{Top Row (Left to Right)}: Simulation environments circuit, circuit2 and levine\_world. The green dot/bulls eye represent the target the robot has to reach. \textbf{Bottom Row (Left to Right)}: Training curves for circuit, circuit2 and levine\_world. The performance of memory augmented actor critic with auxiliary rewards (MA+AC+AR) is compared against the performance of just the memory augmented actor critic (MA+AC), actor-critic with auxiliary rewards (AC+AR) and actor-critic alone. The results depict that the MA+AC+AR architecture performs well on all environments and learns a policy that gets it to the target location. Dark line represents the mean and shaded line represents the standard deviation. Results plotted here are averaged across 3 training runs.}
\label{fig:envs}
\end{figure*}

We test our proposed architecture and reward shaping formulation in simulation as well as on a real robot. The robot is given a small reward (+1) for staying alive. However, to ensure that the robot does not just spin around endlessly collecting infinite reward, we also add a maximum episode length of 500. The hyperparameters for the auxiliary reward layers and the actor-critic layers are set trivially. We notice from our experiments that the performance of the network is mostly immune to the effect of small hyperparameter tuning. For the VI module, and the Memory module, other than the size of the memory, all other hyperparameters are kept same as in~\cite{macn}. For the real world robot experiment, we train in simulation and transfer the policy to real world where an environment approximately similar to the one in simulation is setup. 

\subsection{Computation Graph}
We adapt the MACN for a non holonomic robot operating in a partially observable environment with a sparse range finder as its only sensor. The range finder outputs a sparse lidar scan (100 points). Without any loss in spatial information, the lidar information from state $s_t$ is reshaped to a $m \times n$ input for the encoder network. The encoder network processes this $m \times n$ input and produces an encoding $\phi_{s_t}$. Additionally, the encoder also outputs a reward map $\bar{R}$ and an approximation for the transition probability $f$. These are used in the VI module, to learn a value map for the current state. In the original VIN paper and our earlier work MACN, one has to provide the robot's current pose to the network, since the input to the network is the entire environment and one is only interested in values of states around the current position of the robot. In our work the robot is only presented with information from its current state, and hence we do not need this information about the robots pose at all times. At any given point the VI module only computes a value map for the current position. To improve learning performance, the value map of the current state is subsampled (max-pooled) to produce a low dimension local value map. The local value maps (which can be used to calculate local policies) are sent to a controller network. The controller network interfaces with the memory through an access module (another network layer) and emits read heads, write heads and access heads. This value map is then passed to the differentiable memory scheme. The DNC controller is a LSTM (hidden units vary according to task) that has access to an external memory. The external memory has 64 slots with word size 8. This is the maximum memory we needed in our experiments. The output from the DNC controller and the memory is concatenated through a dense layer. The output of the dense layer is then sent to two different heads that output the current policy estimate (actor layer) and the value estimate(critic) which is used to calculate the TD error and update both the actor and the critic. An action $a_t$ is sampled from the current policy and is used to move the robot to next state $s_{t+1}$ and collect a reward $r_t$. The new state is sent through the encoder to get the encoding of next state $\phi(s_{t+1})$. $\phi(s_t),\phi(s_{t+1}),r_t,a_t$  are used to compute auxiliary losses and pseudo-rewards. In practice, we train the robot on-policy with 30 step discounted returns that are stored in a replay buffer ($\gamma=0.95$). The entire network architecture is then trained end to end according to the actor critic algorithm ~\cite{sutton1998reinforcement}. Adam~\cite{adam} is the sgd variant used to train the network with learning rate of 0.0001. $\eta_{SP}$ and $\eta_{RP}$ are set to 2. $\lambda_{SP}$ is set to 0.2, $\lambda_{RP} = \lambda_{AP} = 0.1$. $\alpha$ is set to 0.5 and $\beta$ is set to 0.01. 

\subsection{Evaluation in Simulation}
The gym-gazebo simulation stack developed by Zamora et. al ~\cite{gym-gazebo} is used for simulating realistic robot experiments. For simulation, the proposed algorithm is tested in 3 environments; namely circuit, circuit2 and levine\_world (Fig ~\ref{fig:envs} (Top Row)). We establish baselines by running actor-critic (AC), actor-critic with auxiliary rewards (AC + AR), memory augmented actor-critic (MA + AC) against our proposed memory augmented actor-critic with auxiliary rewards (MA + AC + AR) (Fig ~\ref{fig:envs}, Bottom Row). It is important to note that the AC, AC+AR also have a LSTM layer to model temporal dependencies. 
\subsubsection{Circuit World}
The \textit{circuit} environment is a simple environment and is aimed at analyzing the feasibility of the proposed architecture to solve the partially observable target problem. For this problem, we reward the robot with a scalar value of $500$ when it gets to the target. Even for this simple environment, we see that the robot is unable to converge to a policy when using just actor critic with memory. For this experiment, the actor-critic with auxiliary rewards converges to goal reaching policy quickly. The memory augmented actor-critic with auxiliary rewards also converges to the optimal policy but takes longer. This can be attributed to the fact that the memory augmented network requires more parameters to train. 
\subsubsection{Circuit World 2}
The \textit{circuit2} environment is a more complicated environment with long hallways.
The \textit{circuit2} environment is aimed at analyzing the use of memory and auxiliary rewards. In $60\%$ of the training episodes, the bookshelf blocking the robots path is not present. This ensures that the robot must always explore the shortest path first. In the event that its path is blocked, the robot then uses its memory to recall states visited in the past and takes the longer route to get to the goal. For this environment, the robot is rewarded with a scalar value of $1000$ when it gets to the goal. Our proposed algorithm performs better than all other baselines and is able to learn a policy that helps it get to the goal even when the easier path to the goal is blocked. 
\subsubsection{Levine World}
The levine\_world represents a realistic office building with long hallways and obstacles such as chairs and trashcans that the robot must navigate to find a path to the target goal location. In this setup the robot is rewarded with a scalar value of $500$ when it gets to the goal. We observe from our experiments that even for this complex world, our proposed methodology is able to converge to the optimal policy in a very small number of episodes.

 \begin{table}[ht]
\caption{\textbf{Simulation Results} Average rewards for 500 episodes after training for 4000 episodes} 
\centering 
\begin{tabular}{c c c c} 
\hline\hline 
Model & circuit & circuit2 & levine world \\ [0.5ex] 
\hline 
AC & 34 & 200 & 18.75 \\ 
AR+AC & \textbf{508} & 204 & 30.3 \\
MA+AC & -70 & 116 &  103\\
MA+AC+AR & 462 & \textbf{818} & \textbf{513.19} \\
\hline 
\end{tabular}
\label{table:nonlin} 
\end{table}
\subsection{Robot Evaluation}
\begin{figure}%
    \centering
    \subfloat[Scarab Robot]{{\includegraphics[height=3.5cm,width=2.8cm]{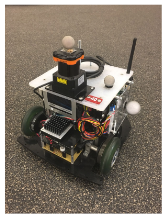} }}%
    \subfloat[Circuit World 2]{{\includegraphics[height = 3.5cm,width=5cm]{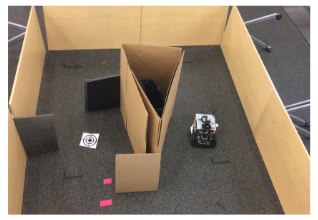} }}%
    \caption{Experimental robot platform and testing environment}%
    \label{fig:robot}%
\end{figure}

To evaluate the performance of the proposed algorithm, we deploy our code on a mobile nonholonomic robot Fig. \ref{fig:robot}(a). An environment approximately similar in scale and size to circuit world is setup using furniture and miscellaneous equipment Fig. \ref{fig:robot}(b). equipped with an  on-board  computer, wireless communication, and a Hokuyo URG laser range finder. It is actuated by stepper motors and its physical dimensions are 30 x 28 x 20 cm with a mass of 8kg. Laser scans were received at a rate of 10Hz and 100 range readings were obtained as in simulation. Angular velocity commands were issued at 10Hz. As before, the robot gets a reward of $500$ when it gets to the goal and a reward of $+1$ for staying alive. Additionally, each episode is constrained to $500$ timesteps. To account for changes in controller dynamics and environment dynamics, we fine tune our policies trained in simulation in the real world. This is done by executing the policy trained in simulation in the real world to collect rewards. Using these new rewards, policy gradients are computed and network parameters are updated. To evaluate performance of different models, we calculate the average reward collected by the reward from the environment over 3 episodes after fine tuning is done. We use a Vicon setup to calculate the robots current pose. This is used only to check if the robot has reached the goal or not and is not fed to the robot or used in any other computation.

We observe that when using the memory augmented actor critic with auxiliary rewards, the robot is able to converge to the goal in under 30 training episodes. This quick convergence in addition to other factors can also be attributed to using the encoder which is able to operate on high level features from the environment. These high level features for a lidar scan are not too different when transferring from simulation to real world. Another interesting result is that AR+AC which had  performed better than MA+AC+AR on circuit world in simulation does not emulate the same performance when transferred to the real world. Link to video performance can be found \href{https://youtu.be/k8CKsAToMak}{here}.

\begin{figure}[t!]
  \centering
  \includegraphics[scale = 0.18]{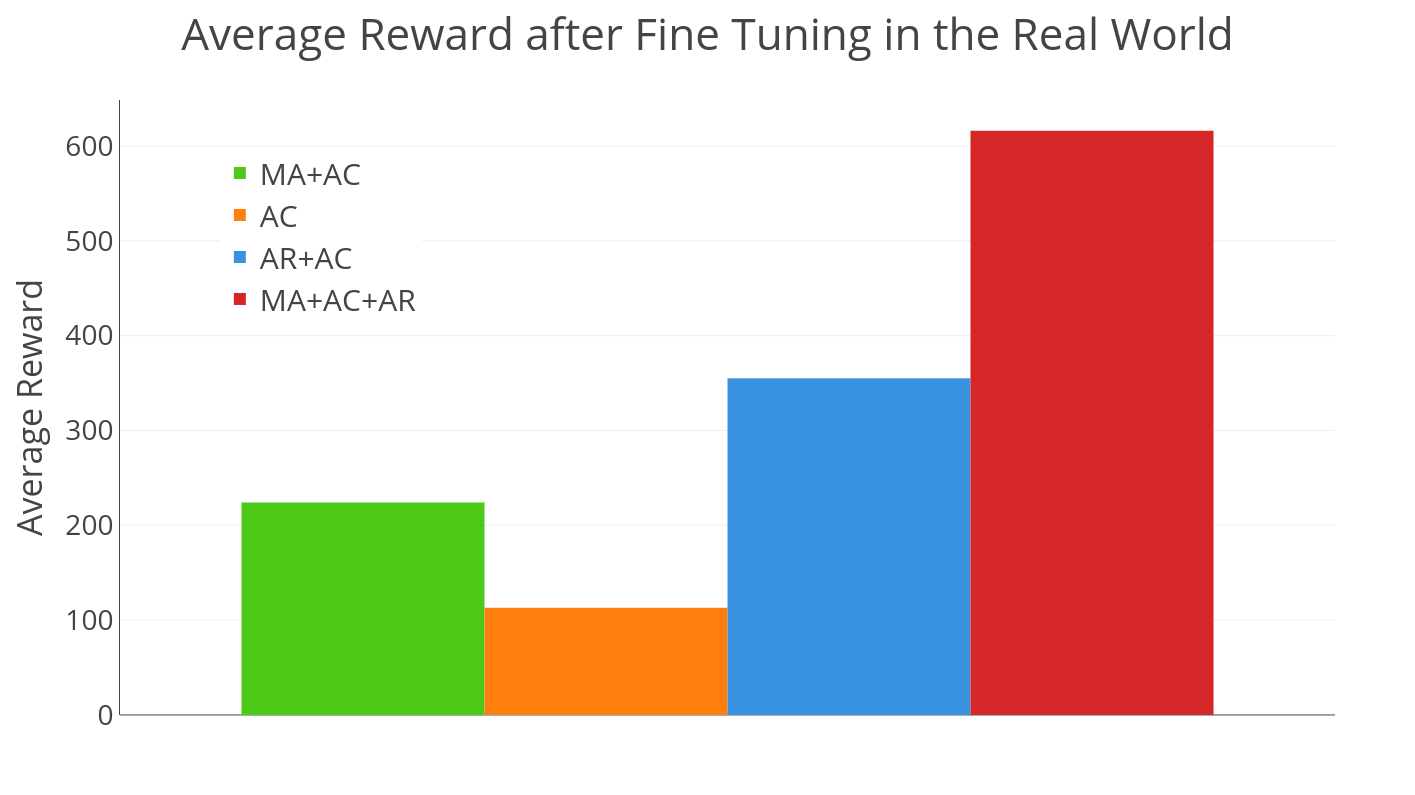}
  \caption{\textbf{Average Reward after Fine Tuning Simulation Policies} After fine tuning for just 30 episodes, only MA+AC+AR is able to converge to the goal. \label{fig:realworldplot}}
\end{figure}

\section{Conclusion}
Deep reinforcement learning has been successful at learning how to solve smaller sub problem s relevant to robotics such as learning how to plan or learning how to explore an unknown environment with sparse rewards. Very few works attempt to solve both problems simultaneously. The MA+AC+AR model learns how to plan and exploits any self-repeating structure in the environment by using explicit external memory. This type of memory is more powerful than a simple recurrent network, thus making the algorithm converge faster. Further, it is easy to adapt to environments with sparse rewards by extending the planning computation to force exploration. In future work, we intend to adapt this framework to physical robotic tasks such as exploring an environment, or a robot arm that learns to execute a series of complex actions. 

\section*{ACKNOWLEDGMENT}
We gratefully acknowledge the support of
ARL grants  W911NF-08-2-0004 
and W911NF-10-2-0016, 
ARO grant W911NF-13-1-0350, 
N00014-14-1-0510, 
DARPA grant HR001151626/HR0011516850 
and IIS-1328805.

\addtolength{\textheight}{-1cm}   





\bibliographystyle{IEEEtran}
\bibliography{example}

\end{document}